# Warehouse Layout Method Based on Ant Colony and Backtracking Algorithm

Ardy Wibowo Haryanto, Adhi Kusnadi, Yustinus Eko Soelistio
Informatics Engineering Program
Faculty of Information and Communication Technology, Multimedia Nusantara University
Boulevard Raya 26, Gading Serpong, Banten
www.awh.tb1@gmail.com, adhikusnadi@yahoo.com, yustinus.eko@umn.ac.id

*Abstract*—Warehouse is one of the important aspects of a company. Therefore, it is necessary to improve Warehouse Management System (WMS) to have a simple function that can determine the layout of the storage goods. In this paper we propose an improved warehouse layout method based on ant colony algorithm and backtracking algorithm. The method works on two steps. First, it generates a solutions parameter tree from backtracking algorithm. Then second, it deducts the solutions parameter by using a combination of ant colony algorithm and backtracking algorithm. This method was tested by measuring the time needed to build the tree and to fill up the space using two scenarios. The method needs 0.294 to 33.15 seconds to construct the tree and 3.23 seconds (best case) to 61.41 minutes (worst case) to fill up the warehouse. This method is proved to be an attractive alternative solution for warehouse layout system.

*Keywords—warehouse layout; block stacking method; ant colony algorithm; backtracking algorithm.*

## I. PRELIMINARY

Warehouse layout configuration problem is still an active area of research. One study conducted by Ahmad Hambali (2011), proposes a warehouse layout method based on block stacking and ant colony algorithm (ACA).The method works by choosing the best combination that have been produced by a group of ants that have chosen the route in stages based on the value of the function of pheromones and heuristic information best. It can explore all the possibilities that exist and adjust for changes in real time. Unfortunately its application limited by no categorization of goods and the number and size of the goods has been known. Another study conducted by Liliana, Gregorius Satia Budhi, and Arief Abadi (2010), propose a warehouse layout method based on backtracking algorithm (BA). The method work by checking each of the possible solution to get the best solution based on the similarity of category and availability. It can find the best solution however it complex and needs a considerably long processing time.

It is the aim of this paper to suggest another warehouse layout method that based on ACA and BA that can work that is (1) fast, (2) works with multiple goods' categories, (3) provide the flexibility for the number and size of the goods. The outline of this paper is as follows. In section II we convey the studies that have been conducted. Then in section III, we describe the proposed method.

## II. THEORIES AND PREVIOUS STUDIES

### A. Ant Colony Algorithm

ACA can be used as an alternative method to solve the optimization problem of space (Hambali: 2011). ACA is a parameter searching method that mimics the behavior of ant colony to looking for food.ACA works in two steps. In the first step, the search begins with the gradually searching each state until a solution is found. The selection is based on the greatest probability value of the pheromone trail and heuristic information contained in each state that calculated using the formula:

$$P_{ij} = \frac{(T_{ij}^{\alpha})(\eta_{ij}^{\beta})}{\sum (T_{ij}^{\alpha})(\eta_{ij}^{\beta})}$$

where $P_{ij}$ is the probability of choosing a state, $T_{ij}$ is intensity of the pheromone trail for inter-state, $\eta_{ij}$ is visibility of a solution that would selected by ant, α is parameter that controls the intensity of pheromone trail where α ≥ 0, β is parameter that control the visibility where β ≥ 0. Pheromone trail corresponds to the quality of the solutions that have been produced by the ant from its prior movement, while heuristic information corresponds to the data input of a problem. This searching is gradually carried out by all ants in the colony (an ant colony).

The second step, the method select the best solution from a set of probable solutions that generated in the previous step. The searching will be processed by all colony (the process is measured by the number of cycles).The best solution obtained from an ant colony will be compared with the best solution obtained from another ant colony (other cycles). After each cycle, the method recalculate the new pheromone trail in each state using the formula:

$$T_{ij} = (1-\rho)T_{ij} + \rho \Delta T_{ij}$$

where $T_{ij}$ is intensity of the pheromone trail for inter-state andρ is constant evaporation of pheromone trail, 0 > ρ > 1.

### B. Backtracking Algorithm

BA can be used as an alternative solution to solve goods layout problem in a parameter space (Liliana: 2010). Arrangement begins with determining the location of the same goods as the new incoming goods to be placed. If there are goods that have not been placed at this early stage, the next stage will be called in recursion manner. Backtracking is applied when the solution has to be tracked back to its previous steps. BA has the advantages of its ability to obtain the best results from a set of combinations by exhaustively searching in all possible solution space. However, this algorithm is not efficient because the search process could take a long time when the search space is large.

### III. PROPOSED METHOD

The propose warehouse layout method consists of two steps. First, the method constructs a solution tree using BA. Then second, it searches for the best solution using ACA and BA.

*A. TreeConstructionStep*

The tree consists of five levels. The first level is the category of goods in warehouse layout (e.g. food, glassware, electronic. The second level is the storage space which is measured in terms of blocks. The block is a three dimensional storage space (in Euclidean sense) that has length x (third level), height y (fourth level), and width z (fifth level). The size of all levels can predefine. It is common that a binary tree structure is used as a solutions tree, however the binary structure will pose complexity problem when the solutions form an unbalance tree. Therefore we propose to use a tree's structure that each node has exactly five child nodes.

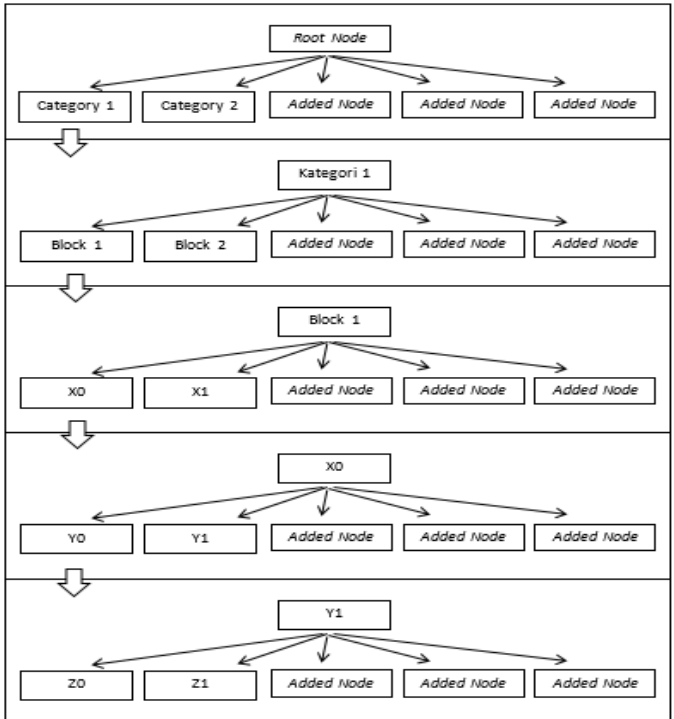

Fig. 1.  Five levels of tree

Each node in the tree stores the (1) name of the node, (2) node binary status (0 = empty, 1 = full), (3) intensity of pheromone trail for inter-state($T_{ij}$),(4) visibility of a solution that would selected by ant ($\eta_{ij}$), and (5)amount of pheromone trail deposited ($\Delta T_{ij}$).The default parameter for each node are$T_{ij}$, and$\Delta T_{ij}$with a value of zero, and$\eta_{ij}$. The visualisation of the construction step can be seen in figure 3.

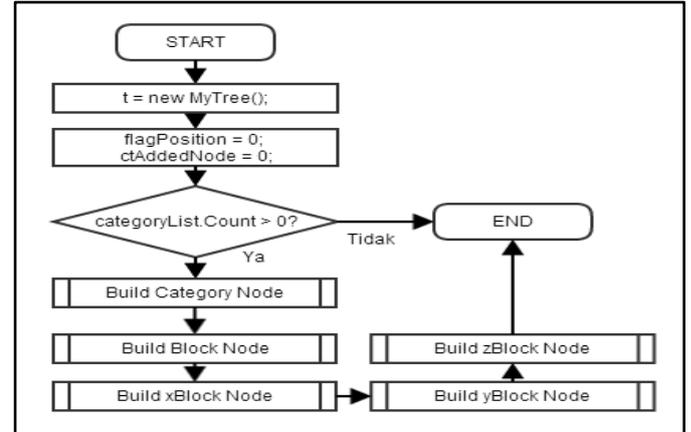

Fig. 2.  Tree construction flowchart

*B. Solution Searching Step*

To search the best fit solution from the tree, this method employ ACA and BA in cascading manners.ACA was first used in the search process and BA as backup when ACA cannot find a fit solution. Each ant will perform a search from the first to fifth level by tracking the amount of pheromone trail and probability exists on the each node. If ACA can not find a fit solution, then BA will be employed with leaf nodes (the solution does not fit the previously obtained) as a starting point.

ACA is used in this method provide the flexibility for the number of goods (the number of goods to be laid out unknown) and has only use one cycle.

All node, except the root, are initialized by using the probability of a number of possibilities existas follows:

- Category Node

$$\eta_{ij_x} = \frac{1}{categoryCount}$$

- Block Node

$$\eta_{ij_x} = \frac{blockVolume_x}{\sum_{x=0}^{x=y} blockVolume_{x^*}}$$

Description,
* = block volume that has the same category.

- Block Length (x) Node

$$\eta_{ij_x} = \frac{1}{blockLength}$$

- Block Height (y) Node

$$\eta_{ij_x} = \frac{1}{BlockHeight}$$

- Block Width (z) Node

$$\eta_{ij_x} = \frac{1}{blockWidth}$$

An incoming goods is regarded as an ant. An ant will move from node i (the parent node) to node j (child nodes)

based on the greatest probability value calculation results based on the pheromone trail and heuristic information contained in each state.

Prior to the movement from current node to one of its child nodes, do validation against the fifth child node whether the fifth child nodes and child nodes of the child nodes, and so on, have an empty space or not. If not, then value of the node and value of $\eta_{ij}$ node from current node changed to zero and go up to the parent node of the current node. By changing the value of $\eta_{ij}$ node from current node, then the probability of election of the nodes become smaller.

For the movement of ants from root node to category node, do addition against $\eta_{ij}$ value of the category node if the goods that entered has the same category with the category nodes. If the goods category are not in accordance with all the category node at first level, then the ants move to the parent node (added nodes) of the node category at second level, and so on.

For the movement of ants from category node to block node, do reduction against $\eta_{ij}$ value of the block node to be zero if the goods that entered has the size (length, width, height) or volume larger than the size or volume of the block nodes. If the size or volume of the goods do not conform to any size or volume block node at first level, then automatically the ant moves to the parent node (added nodes) from block node at second level, and so on. If there is no space at all block nodes that has the same category, then in the end the ants move towards the category node that have different categories from the category of goods that entered. If that happens, then the search process is stopped (space not found).

If the divisor is zero (due $T_{ij}$ all child node is zero), then recalculate the divisor value by giving $T_{ij}$ value for all child nodes with a default value of $T_{ij}$ established, namely 0.1.

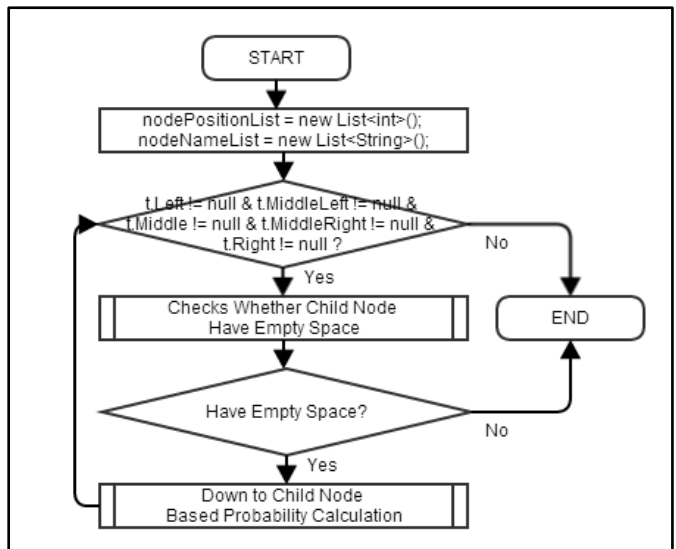

Fig. 3. Ant colony algorithm flowchart

If found enough free space to put a goods, then the intensity of the pheromone trail for inter-state ($T_{ij}$) and the amount of pheromone trail deposited ($\Delta T_{ij}$) for every edge that is passed from the root node to a leaf node until to find a leaf node to be updated. In addition, update value with a value of one and visibility of a solution that would be chosen by ($\eta_{ij}$) with a value of zero for all *leaf nodes* that describe the smallest unit of storage volume has been filled goods.

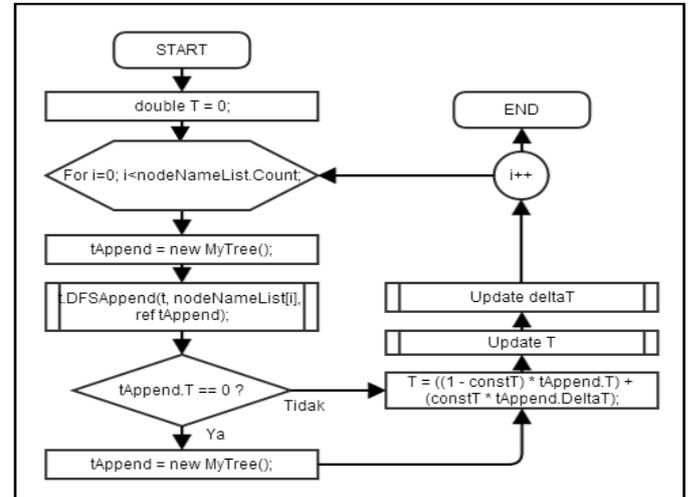

Fig. 4. Ant colony algorithm flowchart (*global pheromone trail update*)

## C. Backtracking Algorithm Flowchart

If ACA can not find a fit solution, then this method use BA as alternative with leaf nodes (the solution does not fit the previously obtained) as a starting point. Backtracking is done until obtain space to put the pieces of the goods or all nodes that are in the same category have visited (there is not enough free space to put a goods).

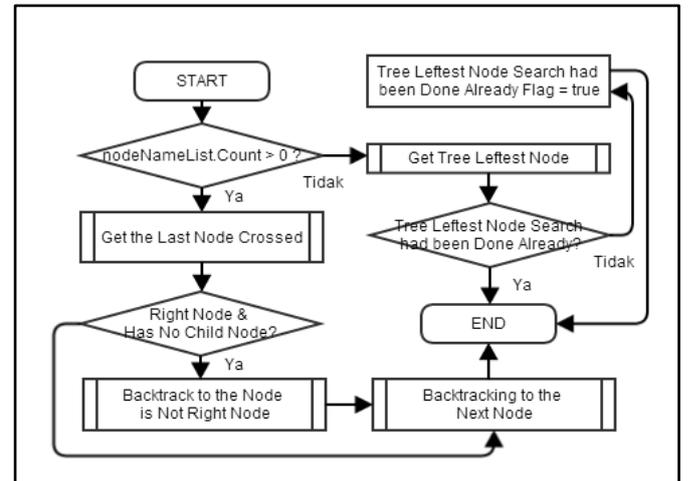

Fig. 5. Backtracking algorithm flowchart

## IV. TEST AND COMPARISON

The method is tested using two scenarios. The first scenario test the method using homogeneous goods size as input set. The second scenario test the method using heterogeneous goods size as input set.

In the first scenario, the number of category is determine with $1 \leq n_c \leq 2$. Each category has the number of block which determined with $1 \leq n_b \leq 3$. The size of each block is determine with $1 \leq \{x_b, y_b, z_b\} \leq 6$. The size of goods is determine with $1 \leq \{x_g, y_g, z_g\} \leq 6$. The method run until goods can no longer be accommodated. To finish the first scenario, the method needs 2.294 seconds to 54.19 minutes with 100% space occupation.

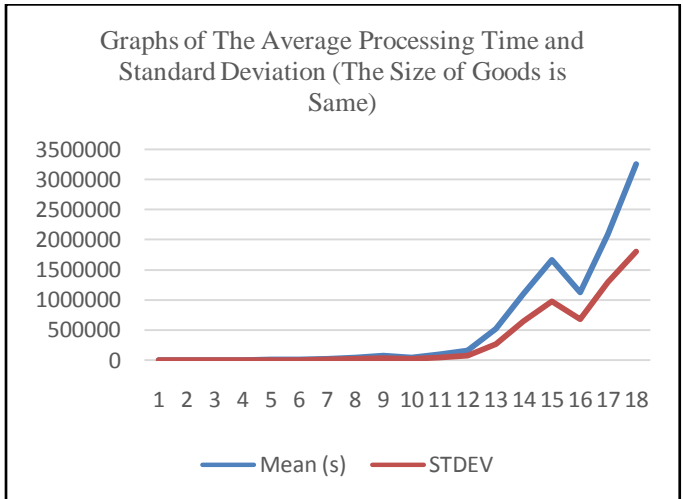

Fig. 6. Graphs of The Average Processing Time and Standard Deviation (The Size of Goods is Same)

The x-axis with index 1 to 18 describe the increase in the size of the space. The y-axis describe time process. An increase in the size of the space is influenced by increasing the number of categories, the number of blocks, and the size of blocks. From Fig. 6., it can be seen that the increase in the size of the space causes a greater time process for process in the arrangement of goods that have the same size

In the second scenario, the number of category is determine with $1 \leq n_c \leq 2$. Each category has the number of block which determined with $1 \leq n_b \leq 3$. The size of each block is determine with $1 \leq \{x_b, y_b, z_b\} \leq 6$. The size of goods is $x_g$ = random between 1 until $1/3 x_b$, $y_g$ = random between 1 until $1/3 y_b$, $z_g$ = random between 1 until $1/3 z_b$. In this scenario, the method needs 3.225 seconds to 61.41 minutes to be completed. The occupation from this scenario is 97.19% to 100%.

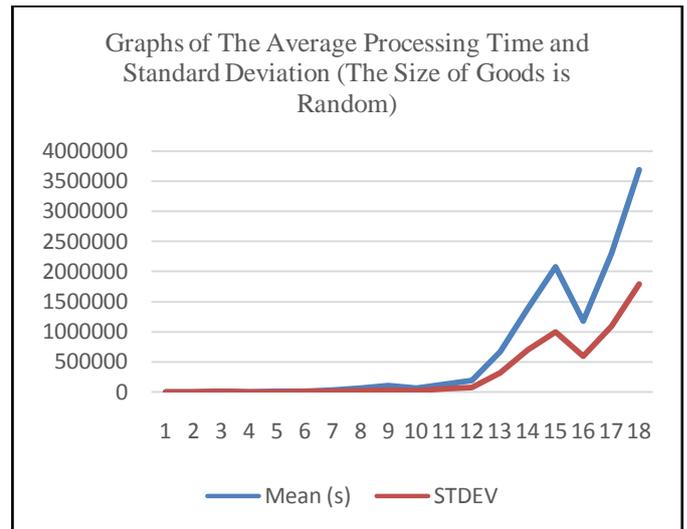

Fig. 7. Graphs of The Average Processing Time and Standard Deviation (The Size of Goods is Random)

From Fig. 7., it can be seen that the increase in the size of the space causes a greater time process for process in the arrangement of goods that have the random size. The arrangement of goods with the same size and the arrangement of the goods with random sizes have a similar time process which indicates that this method can be used to arrange the goods that have the same size and different sizes.

V. DISCUSSION

The method has been tested and shows a promising result. The test from both scenarios show that the method has good level of speed and occupation, works with multiple goods' categories, and provide the flexibility for the number and size of the goods.

However, the performance of this method is still limited due to the size of the tree that greatly influence the performance of search. Its usability is also still limited only for block and goods with square shape.

VI. CONCLUSION

A new alternative warehouse layout approach based on ACA, BA, and block stacking method successfully applied in the goods layout application. This method combines the use of BA and ACA, which in previous studies ACA used to organize a number of goods that have been known and BA used to manage the goods where it is possible to have different categories (however it is complex and needs a considerably long processing time). It is fast, works with multiple goods' categories, and provides the flexibility for the number and size of the goods.

For future development, exploring a new tree structure or a data structure can be can be done to increase the processing speed. The application can be modified to handle the block and goods in addition to the form of cubes and blocks.